\documentclass[10pt,journal,compsoc]{IEEEtran}
\ifCLASSOPTIONcompsoc
  \usepackage[nocompress]{cite}
\else
  \usepackage{cite}
\fi
\ifCLASSINFOpdf
\else
\fi
\usepackage{lipsum}
\usepackage{graphicx}
\usepackage{booktabs}
\usepackage{amsmath} 
\usepackage{tcolorbox}
\usepackage{multirow}
\usepackage{algorithm}
\usepackage{algpseudocode}
\usepackage{url} 

\begin{document}
%
\title{A Self-feedback Knowledge Elicitation Approach for Chemical Reaction Predictions}
%
%
%
%

\author{Pengfei~Liu,
        Jun~Tao,
        and~Zhixiang~Ren
\IEEEcompsocitemizethanks{
\IEEEcompsocthanksitem Pengfei Liu is with the School of Computer Science and Engineering, Sun Yat-sen University, and Peng Cheng Laboratory. 
\IEEEcompsocthanksitem Jun Tao is with the School of Computer Science and Engineering, Sun Yat-sen University. 
\IEEEcompsocthanksitem Zhixiang Ren\textsuperscript{*} is with Peng Cheng Laboratory. E-mail: jason.zhixiang.ren@outlook.com
}
\thanks{\textsuperscript{*}Corresponding author.}
}

%
%

\markboth{Journal of \LaTeX\ Class Files,~Vol.~XX, No.~XX, XX~2024}%
{Shell \MakeLowercase{\textit{et al.}}:}
%



\IEEEtitleabstractindextext{%
\begin{abstract}
The task of chemical reaction predictions (CRPs) plays a pivotal role in advancing drug discovery and material science. 
However, its effectiveness is constrained by the vast and uncertain chemical reaction space and challenges in capturing reaction selectivity, particularly due to existing methods' limitations in exploiting the data's inherent knowledge.
To address these challenges, we introduce a data-curated self-feedback knowledge elicitation approach.
This method starts from iterative optimization of molecular representations and facilitates the extraction of knowledge on chemical reaction types (RTs).
Then, we employ adaptive prompt learning to infuse the prior knowledge into the large language model (LLM).
As a result, we achieve significant enhancements: a 14.2\% increase in retrosynthesis prediction accuracy, a 74.2\% rise in reagent prediction accuracy, and an expansion in the model's capability for handling multi-task chemical reactions.
This research offers a novel paradigm for knowledge elicitation in scientific research and showcases the untapped potential of LLMs in CRPs.
\end{abstract}

}

\maketitle

\IEEEdisplaynontitleabstractindextext

%
\IEEEpeerreviewmaketitle

\IEEEraisesectionheading{\section{Introduction}\label{sec:intro}}

%
%
%
%
\IEEEPARstart{T}{he} applications of CRPs \cite{shilpa2023recent} span drug discovery, material science, and synthetic pathway optimization, which have a critical role in advancing various scientific fields.
The primary challenges within this domain arise from the vast and uncertain chemical reaction space, coupled with the complexities of reaction selectivity.
Moreover, most existing methods fail to harness the intrinsic knowledge within reaction data.
Traditional methods in CRPs \cite{jorgensen1990cameo}\cite{hollering2000simulation} struggle to navigate the intricate and variable dynamics of chemical reactions, which rely on domain expertise and heuristic strategies.
While existing computational methods have strived to predict chemical reactions, they often fall short in handling the inherent complexity and selectivity due to limited datasets and the lack of detailed reaction mechanism guidance.
With advancements in artificial intelligence (AI), AI methods \cite{kayala2012reactionpredictor}\cite{wei2016neural}\cite{do2019graph} have led to notable improvements in the accuracy of CRPs.
However, the challenge of generalizing these improvements across diverse chemical reactions persists.

\begin{figure*}[!htb]
\centering
\includegraphics[width=0.95\textwidth]{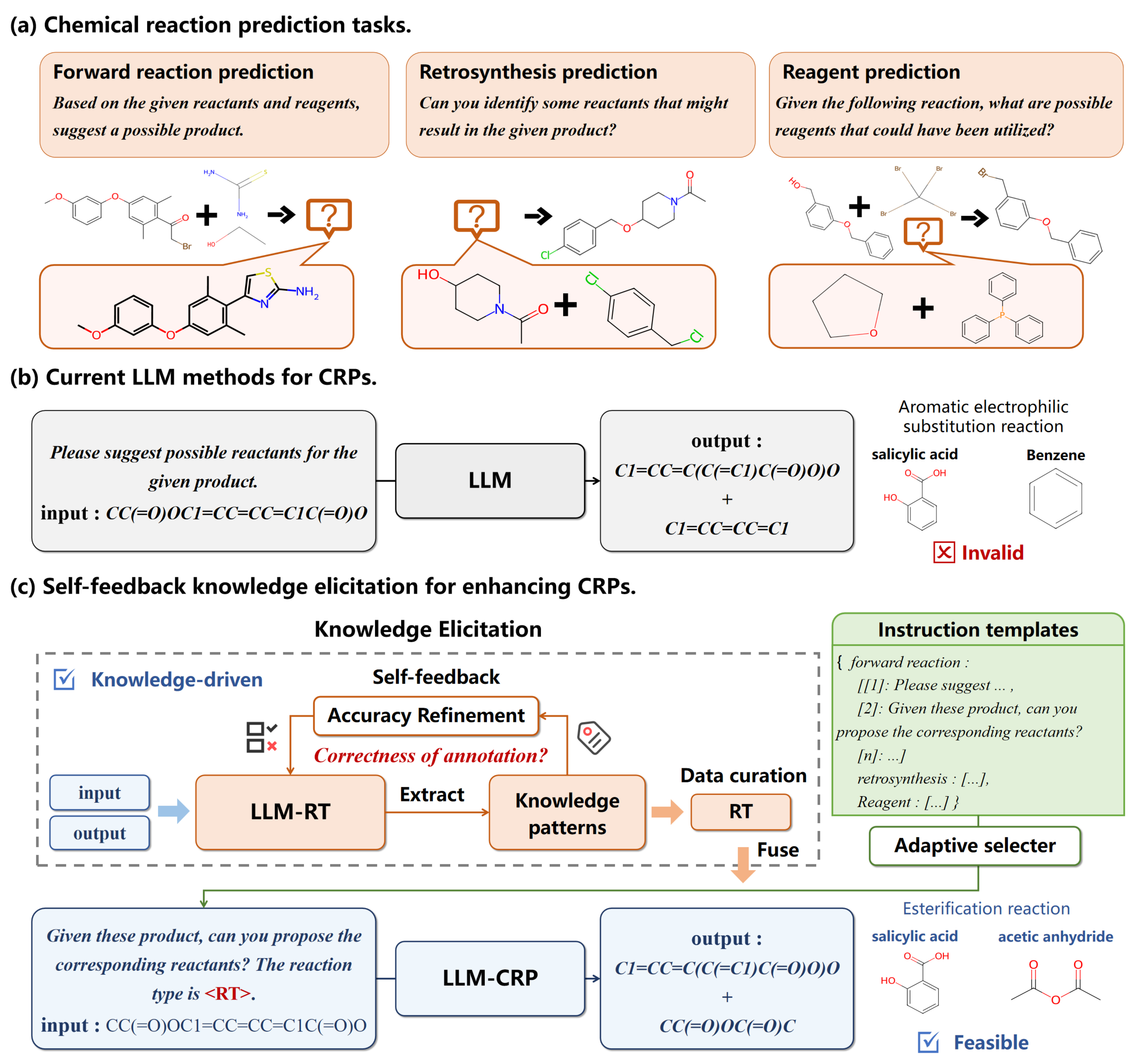}
\caption{\textbf{Overview of tasks and approaches}. \textbf{(a) Chemical reaction prediction tasks}, showcasing three tasks along with examples.
\textbf{(b) Current LLM methods for CRPs}, indicating rational predictions but lacking in reactive validity.
\textbf{(c) Self-feedback knowledge elicitation for enhancing CRPs}, demonstrating the enhancement of CRPs through the refinement of knowledge patterns, notably RTs, utilizing a self-feedback knowledge elicitation technique.
Knowledge elicitation serves as a method of data curation for knowledge distillation, where RT is integrated into large language models via adaptive prompt learning, facilitating the planning of reaction pathways in CRPs.
}
\label{fig: overview}
\end{figure*}

With the emergence of ChatGPT \cite{schulman2022chatgpt} and GPT-4 \cite{achiam2023gpt}, LLMs have gained attention for their potential in various domains, including science.
LLMs contribute a new trend in scientific language modeling (SLM) \cite{liu2024scientific}, with models like Galactica \cite{taylor2022galactica} focusing on the scientific domain.
Smaller pre-trained models such as MolT5 \cite{edwards2022translation} and BioT5 \cite{pei2023biot5} have started to be applied in molecular SLM tasks.
Similarly, Text+Chem T5 \cite{christofidellis2023unifying} and InstructMol \cite{cao2023instructmol} cover CRP tasks, yet large models still face challenges in interpretability and the demand for extensive training data.

Prompt learning, coupled with LLMs for fine-tuning, has emerged as a standard paradigm, offering a way to integrate domain-specific knowledge into model training \cite{fang2023mol}.
However, static template prompts can lead to rigid guiding patterns in LLMs, potentially impacting their generalizability.
The dynamic prompts can tackle the limitations of static templates with the injection of prior knowledge into LLMs. 
In CRPs, identifying RTs can narrow down the chemical space to be explored.
However, the datasets typically lack RT labels, including only reactants and products.
While too few categories in annotation methods limit their effectiveness, too many can reduce annotation accuracy.
Meanwhile, the synergistic effect of multi-task cooperative learning, treating chemical reactions as a unified domain of molecular knowledge, has yet to be fully leveraged.
Therefore, we can conclude several key issues:
\textbf{(1) How can we balance annotation accuracy and number of RTs in knowledge elicitation by LLMs?}
\textbf{(2) Can LLMs perform better through prompt-based knowledge infusion?}
\textbf{(3) Can a multi-task collaborative approach improve the performance of LLMs?}

To address these issues, we introduce a prompt-based knowledge elicitation \cite{xu2024survey} approach that combines knowledge distillation and integration through adaptive prompts, aiming to boost the accuracy of LLMs.
The task of CRP is broken down into RT prediction and molecule generation.
By applying a self-feedback knowledge elicitation method for high-accuracy annotating RTs and utilizing prompt learning for knowledge infusion into the LLM, we enhance model performance and achieve the synergistic benefits of multi-tasking.

In summary, our main contributions are the following:
\begin{itemize}
    \item \textbf{Self-Feedback Knowledge Elicitation:} We propose a novel knowledge elicitation approach by integrating a self-feedback mechanism with data curation using LLM, enhancing accuracy in CRPs.
    \item \textbf{Dynamic Prompting for LLMs:} We introduce a dynamic prompt learning to address the limitations of static prompts, achieving a 10\% additional increase in the knowledge injection adaptability of LLMs.
    \item \textbf{Synergistic Multi-Task Enhancement:} By injecting prior knowledge, we facilitate a synergistic improvement of 14.9\% across reaction prediction tasks.
\end{itemize}

The rest of this paper is organized as follows. 
Section~\ref{sec:rel} offers a review of existing studies, laying the groundwork for our approach.
In Section~\ref{sec:met}, we delve into the methodology, detailing the data, models, and strategies used.
Section~\ref{sec:eva} is dedicated to evaluating our methods and includes an ablation study to highlight key findings.
The discussion in Section~\ref{sec:res} centers on our annotation approach, the knowledge-learning capabilities of large models, and the benefits of multi-tasking.
Finally, Section~\ref{sec:con} summarizes our main contributions and suggests future research directions.
The data and software can be accessed at \url{https://github.com/AI-HPC-Research-Team/SLM4CRP}.

\section{Related Work}
\label{sec:rel}
In this section, we focus on the models for CRPs, the application of LLMs, and the innovative use of prompt learning and knowledge priors \cite{braun2022exact}, laying the groundwork for our proposed approach.

\subsection{Chemical Reaction Predictions}

As illustrated in Figure \ref{fig: overview} (a), CRP tasks involve determining the products or reactants of chemical reactions from given molecules.
The key to the CRPs lies in accurately identifying the mechanisms and outcomes involving bond breakage and reformation under a variety of conditions.
It makes the predictions of reactions particularly complex due to the vast number of possible reaction mechanisms and products.
It requires a deep understanding of chemical knowledge and molecular interactions to forecast the most probable pathways for forward reactions, retrosynthetic routes, and necessary reagents for specific transformations.

For traditional methods, CAMEO \cite{jorgensen1990cameo} leverages detailed heuristics across chemical classes to predict multistep reactions, while EROS \cite{hollering2000simulation} utilizes a graph-based rule library enhanced by additional constraints from physical data and kinetic simulations.  Despite their sophistication, these traditional methods often struggle with the vast complexity and variability of chemical reactions, leading to limitations in predictive accuracy and scalability. 

With the development of AI methods, the ReactionPredictor \cite{kayala2012reactionpredictor} model narrows down the reaction space through a filtering model and then ranks the prioritized likely reactions.
Another method \cite{wei2016neural} combines fingerprints of reactants and reagents into a reaction fingerprint, used as input to a neural network predicting probabilities across 17 RTs.
Additionally, the GTPN \cite{do2019graph} model leverages graph neural networks \cite{wang2023graph} (GNN) to comprehend the molecular graph structures of input reactants and reagents, refining the prediction of correct products through reinforcement learning.
The Molecular Transformer \cite{schwaller2019molecular}, based on Transformer \cite{vaswani2017attention}, treats reaction prediction as a machine translation issue between molecules represented in SMILES (Simplified Molecular Input Line Entry System) \cite{weininger1988smiles} strings, further enabling the assessment of prediction uncertainty.

Traditional methods in CRPs rely on heuristics and rule-based systems, often limiting their adaptability and scalability across the diverse landscape of chemical reactions.
As for AI approaches, despite their innovative frameworks and predictive power, their challenges include data dependency, model interpretability, and the generalization of predictions to unseen reactions.

\subsection{Knowledge Distillation and Elicitation}

Knowledge distillation \cite{gou2021knowledge} is a machine learning technique in which knowledge is transferred from a larger, more complex model (teacher) to a smaller, simpler model (student).
This method is commonly employed for model compression \cite{choudhary2020comprehensive}, aiming to enhance model performance while adhering to a fixed capacity constraint.
It allows the compact student model to mimic the behavior of the larger teacher model, thereby achieving efficiency without significantly compromising the quality of the model.
The survey \cite{xu2024survey} categorizes the pipeline of distilling knowledge from LLMs into two main phases: knowledge elicitation and the distillation algorithm.
Moreover, they identify six methods of knowledge elicitation from teacher LLMs.
Among these, the data curation approach has gained attention for its focus on producing high-quality and scalable data generation for knowledge distillation purposes.
Unlike data augmentation \cite{liu2020survey}, which primarily aims at increasing the quantity of training data, data curation emphasizes constructing a high-quality training dataset.

For instance, InPars \cite{bonifacio2022inpars} leverages the LLM to generate labeled data in a few-shot manner, creating synthetic datasets that enhance performance in information retrieval tasks.
Similarly, ZEROGEN \cite{ye2022zerogen} employs LLMs to generate unsupervised datasets, upon which a smaller task-specific model is trained, facilitating efficient inference across various Natural Language Processing (NLP) tasks.
These data curation techniques can act as variants of knowledge distillation, producing high-quality datasets for CRPs and other domains, thereby enriching the pool of strategies for effective knowledge distillation.

\subsection{Large Language Models}

Transformer-based models such as BERT \cite{devlin2018bert}, GPT \cite{radford2019language} and T5 \cite{raffel2020exploring}, showcasing remarkable capabilities in understanding and generating human language.
As the scale of models has grown with data availability, models like Chinchilla \cite{hoffmann2022training}, LLaMA \cite{touvron2023llama}, and GLM \cite{zeng2022glm}, which demonstrate an enhanced capacity to process and generate text.

In specific fields, models are increasingly tailored to domain knowledge learning.
In molecular science, MolT5 \cite{edwards2022translation}, based on the T5 architecture, pioneers tasks in molecular description and text-based molecular design.
Text+Chem T5 \cite{christofidellis2023unifying} represents a multi-task language model approach capable of handling a variety of tasks across both chemical and linguistic domains.
GIT-Mol \cite{liu2024git} introduces an innovative approach by aligning and integrating molecular text, graphs, and images through cross-attention mechanisms and contrastive learning.
Meanwhile, BioT5 \cite{pei2023biot5} merges knowledge from the molecular and protein domains, presenting a cross-disciplinary pre-trained model that underscores the potential of LLMs to bridge and enhance research across fields.
The MOLGEN \cite{fang2023domain} model, built on the BART \cite {lewis2020bart} and utilizing SELFIES \cite{krenn2020self}, is capable of generating novel molecules and optimizing molecular structures based on desired properties.

Although LLMs display enhanced generalization capabilities and a stronger understanding of knowledge, as shown in Figure \ref{fig: overview}(b), they may select incorrect synthetic pathways in CRPs.
Despite these advancements, they still face challenges with data scarcity in specialized domains and lack interpretability.

\begin{figure*}[!t]
\centering
\includegraphics[width=0.95\textwidth]{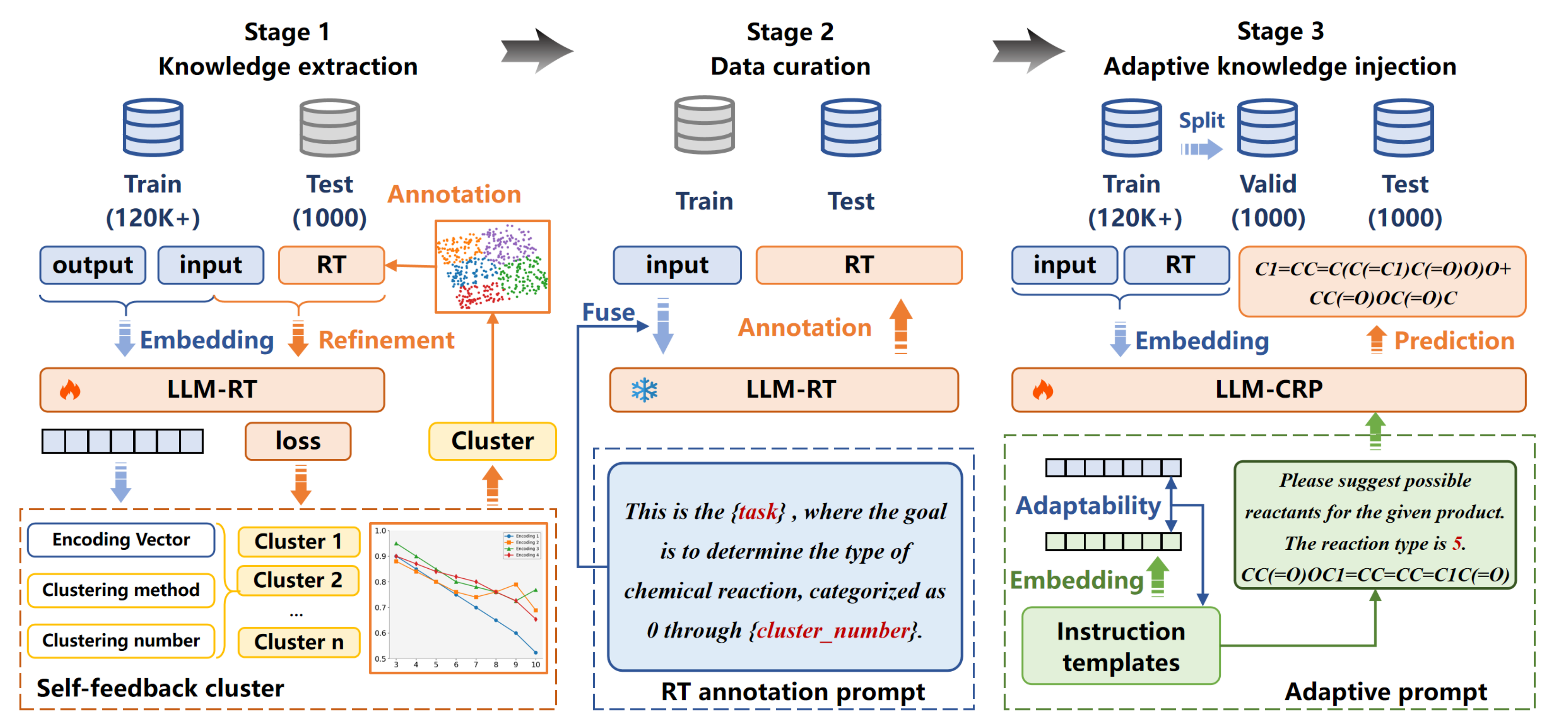}
\caption{\textbf{Three-stage training scheme of prompt-based knowledge elicitation}.
\textbf{Knowledge extraction}, the datasets are divided into train, valid, and test sets.
The training dataset's inputs and outputs are clustered using LLM-RT embeddings, leading to RT annotations. The annotation accuracy of LLM-RT is refined by iteratively tuning cluster parameters and training with input and RT, aiming to improve precision and identify the best cluster.
\textbf{Data curation}, the trained LLM-RT annotates the RTs for the validation and testing datasets based on their inputs.
\textbf{Adaptive knowledge injection}, adaptability is calculated based on the embeddings of inputs and instructions, leading to the selection of adaptive instructions.
It is followed by fine-tuning the LLM with prompts that are enhanced with prior knowledge.}
\label{fig: framework}
\end{figure*}

\subsection{Prompt-based Knowledge Priors}

Prompt learning involves designing input `prompts' that guide the LLMs to perform specific tasks or generate certain types of responses.
This technique leverages the knowledge in pre-trained models, enabling them to apply their understanding of language to new tasks without extensive retraining.
The Mol-Instructions \cite{fang2023mol} dataset facilitates LLM fine-tuning with diverse molecule and protein instructions, including the CRPs datasets of USPTO \cite{wei2010novel} and USPTO\_500MT \cite{lu2022unified}.
InstructMol \cite{cao2023instructmol}, a multi-modal LLM, employs instruction tuning to correlate molecular structures with textual data, using a dual-phase training approach that smartly leverages limited domain-specific datasets for molecule captioning and CRPs.
Following BioT5, BioT5+ \cite{pei2024biot5+} extends training with International Union of Pure and Applied Chemistry (IUPAC) \cite{long1983limit} names of molecules and broadens its application to additional tasks.

Knowledge priors refer to pre-existing, domain-specific knowledge that can be integrated into AI models to enhance their understanding and prediction capabilities \cite{kuang2024impact
}.
The knowledge can be fused with the model architecture, training frameworks, or training data to enrich the model's insights and improve its outcomes.
In molecular science, knowledge priors, such as details on molecular structures, chemical properties, and reaction mechanisms, significantly boost AI models' predictive accuracy.
The KPGT \cite{li2023knowledge} framework exemplifies this by utilizing a graph transformer for molecular graphs with a knowledge-guided pre-training strategy, aiming to understand molecules' structural and semantic knowledge.
Similarly, PGMG \cite{zhu2023pharmacophore}, a pharmacophore-guided deep learning method, innovatively tackles the mapping challenges between pharmacophores and molecules, thereby increasing the diversity of biologically active molecules generated.
For prompt-enhanced method, the KANO \cite{fang2023knowledge} method introduces a chemical element-oriented knowledge graph to encapsulate fundamental knowledge of elements and functional groups.
It employs this graph in a novel molecular contrastive learning approach with functional prompts, effectively leveraging deep domain knowledge throughout the pre-training and fine-tuning stages.

Prompt-based knowledge priors represent an evolution in leveraging domain-specific knowledge within AI models.
By crafting prompts that encapsulate molecular knowledge, researchers can direct the focus of LLMs toward SLM problems.
This strategy not only heightens the accuracy and pertinence of model predictions but also streamlines the integration of intricate scientific knowledge, reducing the dependency on voluminous training data and augmenting the model's interpretability.

\section{Methodology}
\label{sec:met}

In this chapter, we present an overview of our approach, which leverages chemical knowledge through prompt learning to enhance the accuracy of CRPs.
First, we introduce the foundational data structure and the preparatory steps.
Next, we delve into the self-feedback knowledge elicitation process, a pivotal mechanism to unearth knowledge patterns.
Finally, we describe how we train our LLM on specific reaction prediction tasks.

\subsection{Overview}

The approach depicted in Figure \ref{fig: framework} unfolds through a three-stage training strategy.
Initially, the dataset is divided into training, validation, and testing sets, with an emphasis on knowledge extraction to facilitate RT annotation.
This stage involves iteratively refining the selection of the optimal clustering approach and training the LLM-RT, leading to the formation of a self-feedback clustering mechanism.
In the data curation phase, the frozen LLM-RT employs prompts alongside inputs to perform RT annotation on the validation and testing sets.
Finally, the method incorporates the \( prompt_{enhanced} \) for fine-tuning the LLM.

The CRP tasks are classified into three primary categories: forward reactions, reverse reactions, and reagent predictions. Each category can be represented mathematically as follows:

\begin{itemize}
    \item \textbf{Forward Reaction Prediction} aims to predict the products (\(\mathcal{P}\)) for a given set of reactants (\(\mathcal{R}\)), formulated as \(f_{forward}: \mathcal{R} \rightarrow \mathcal{P}\).
    \item \textbf{Retrosynthesis Prediction} seeks to identify potential reactants (\(\mathcal{R}\)) from known products (\(\mathcal{P}\)), represented as \(f_{retrosynthesis}: \mathcal{P} \rightarrow \mathcal{R}\).
    \item \textbf{Reagent Prediction} focuses on determining the reagents (\(\mathcal{G}\)) required for the conversion of reactants to products, expressed as \(f_{reagent}: (\mathcal{R}, \mathcal{P}) \rightarrow \mathcal{G}\).
\end{itemize}
This scenario can be represented by a more generalized function:
\begin{equation}
f_{general}: (\mathcal{R}, \mathcal{P}) \rightarrow (\mathcal{R}', \mathcal{P}', \mathcal{G})
\end{equation}
where \(\mathcal{R}\) and \(\mathcal{P}\) are the sets of reactants and products, respectively, and the function aims to predict \(\mathcal{R}'\) (reactants), \(\mathcal{P}'\) (products), and \(\mathcal{G}\) (reagents) for any given chemical reaction.
Let us define the knowledge-driven prompt, denoted as \( prompt_{enhanced} \), as the combination of adaptive instructions ($IA$) and knowledge priors RTs (\(\mathcal{K}\)):
\begin{equation} 
prompt_{enhanced} = IA + \mathcal{K} 
\end{equation}
Incorporating prior knowledge by \(prompt_{enhanced}\) enhances the specificity and accuracy of the predictions, modifying the general function to:
\begin{equation} 
f_{enhanced}: (\mathcal{R}, \mathcal{P}, prompt_{enhanced})  \rightarrow (\mathcal{R}', \mathcal{P}', \mathcal{G})
\end{equation}
In this function, the \( prompt_{enhanced} \) serves to guide the prediction process by incorporating both adaptive instructions and the specified knowledge priors, leading to the prediction of reactants \(\mathcal{R}'\), products \(\mathcal{P}'\), and the necessary reagents \(\mathcal{G}\).

Distinct from existing methods, our approach integrates RT knowledge priors with LLMs through adaptive prompt learning and self-feedback knowledge elicitation techniques.
It addresses the scarcity of RT information in real-world datasets, and the dynamic prompts prevent rigid pattern guiding in LLMs, offering a solution to enhancing prediction accuracy.

\subsection{Data}

In this study, we utilize the `Molecule-oriented instructions' from the Mol-Instructions \cite{fang2023mol} dataset, which encompasses three chemical reaction tasks. Our data preparation involves converting the `input' into SMILES format, resulting in dataset \(D\). This dataset's `instruction' components are sorted by different task types and integrated into our instruction template library, aiding the infusion of domain-specific knowledge into our models.

\textbf{Data preprocessing}:
Before conducting our experiments, we engage in a thorough data preprocessing regimen to safeguard the dataset's integrity and uniformity.
The primary steps encompass transforming all molecular representations into SMILES format to standardize the molecular data, thereby ensuring compatibility across our experiments.
We rigorously clean the data by removing entries that fail the SMILES conversion process, thus mitigating potential inconsistencies and errors in later analysis stages.

\textbf{Data split}:
We follow the original test set partitioning scheme of the dataset. However, in the `knowledge extraction' stage in Figure \ref{fig: framework}, the training subset \(D_{train}\) is randomly divided in a 98:1:1 ratio to obtain \(D'_{train}\), \(D'_{valid}\), and \(D'_{test}\).
This division facilitates the training of the LLM-RT predictor, which subsequently annotates the testing set \(D_{test}\).
In the `Adaptive knowledge injection' stage, the validation set \(D_{valid}\) is split from \(D_{train}\).

\subsection{Knowledge Extraction and Data Curation}

This subsection explains the process of self-feedback knowledge elicitation in CRP problems and describes the method of data curation for instruction-tuning dataset.

\textbf{Knowledge annotation}
Knowledge extraction from the LLM-RT begins by embedding inputs and outputs of the training dataset $D_{train}$, followed by clustering into corresponding clusters, which are annotated as the RTs.
For the vector encoding of input and output embeddings, we consider four alternative methods: directly using the output vector ($output_{vec}$), subtracting input vector from output vector ($output_{vec}$ - $input_{vec}$), concatenating input and output into a vector ($concat(input_{vec}, output_{vec}$), and the dot product of output and input vectors ($output_{vec}\cdot input_{vec}$).

The selection of our clustering method and number is crucial to balance the accuracy and diversity of RT annotations.
As the number of clusters increases, accurately annotating $D_{test}$ becomes more challenging.
There are several common fundamental types of reactions, including but not limited to synthesis reactions, decomposition reactions, single-replacement reactions, and double-replacement reactions.
While these basic categories can be subdivided, over-detailed classification may not be conducive to generalization and efficiency for an effective prediction model.
Moreover, as the number of clusters increases, the accuracy of unsupervised labeling is likely to decrease.
Furthermore, a high number of clusters may complicate the interpretability of the model, making it harder for users to understand and trust the model's predictions.
Considering these factors, a recommended number of clusters would be between 3 to 12 \cite{2019Types}.
This range should adequately cover the main types of chemical reactions while avoiding the pitfalls of over-segmentation, such as increased model complexity or reduced generalizability. 
Furthermore, we opt for the k-means \cite{ahmed2020k} algorithm as our clustering method, given its efficiency and effectiveness in grouping data into cohesive, distinct clusters that reflect underlying patterns within the RTs.
\begin{equation} 
E_{n} = Embedding(input_{D_{train}}, output_{D_{train}}) 
\end{equation}
\begin{equation} 
R_{n} = Cluster(E_{n})
\end{equation}
\begin{equation} 
Acc_{n} = ST(R_{n}, input_{D'_{train}}, input_{D'_{valid}}, input_{D'_{test}})
\end{equation}
The annotation results $R_{n}$ in the $n^{th}$ iteration, where the training dataset $D_{train,n}$ is encoded and clustered.
$E_n$ denotes the encoding function by LLM-RT, and $Cluster$ represents the clustering operation.
The annotation accuracy $Acc_{n}$ achieved after conducting  the supervised training ($ST$) using the annotation results $R_{n}$ and the training dataset $D_{train,n}$ from the $n^{th}$ iteration.

\begin{algorithm}
\caption{Self-Feedback Knowledge Elicitation Process}
\begin{algorithmic}[1]
\State Initialize the LLM-RT model for encoding methods $encodings$, set cluster method and number range, and prepare instruction templates library.
\For{each encoding $e$ in $encodings$}
    \For{$number=3$ to $12$}
        \State Set cluster number $N = number$, encoding method $E = e$.
        \State Embed inputs and outputs from $D_{train}$ using encoding method $E$.
        \State Perform clustering on the embedded data to identify RTs.
        \State Annotate $D_{train}$ with the identified RTs.
        \State Split $D_{train}$ to $D'_{train}$, $D'_{valid}$ and $D'_{test}$ 
        \State Train the LLM-RT model on $D'$ using the RT annotations.
        \State Evaluate and record the annotation accuracy.
    \EndFor
    \State Evaluate and record the overall best annotation accuracy for encoding $e$.
\EndFor
\State Freeze the LLM-RT model after identifying the optimal $E$ and $N$.
\State Use the optimized LLM-RT model to annotate RTs in $D_{\text{test}}$.
\end{algorithmic}
\end{algorithm}

\textbf{Self-feedback cluster}:
After a round of RT annotation, we conduct supervised training of the LLM-RT model against inputs and RTs, adjusting the model weights to refine annotation accuracy continually.
After a training round, RTs for $D_{train}$ are re-annotated, thus iteratively optimizing the LLM-RT training through a self-feedback mechanism, forming an effective self-feedback clustering.
\begin{equation} 
Optimal(E_{best}, N_{best}) = SF(Acc_{n}, E_{n}, Cluster, N)
\end{equation}
\begin{equation}
{RT}_{D_{train}} = Cluster_{N_{best}}(E_{best})
\end{equation}
The self-feedback (SF) process for selecting the optimal encoding method $E$ and the number of clusters $N$ to maximize the annotation accuracy $Acc_{n}$, and $Optimal$ indicates the chosen optimal encoding method and cluster number after the iteration process concludes.
Then, the $RT_{D_{train}}$ annotated for the training dataset $D_{train}$, derived from clustering the dataset with the best embedding method $E_{best}$ into $N_{best}$ clusters through $Cluster_{N_{best}}$. 
The selection process aims to balance the annotation accuracy and the number of RTs, ensuring optimal categorization.

\begin{tcolorbox}[sharp corners, boxrule=0.5pt]
\textbf{RT annotation prompt:}\\
This is the \textcolor{red}{\{$task$\}} reaction prediction task, where the goal is to determine the type of chemical reaction based on the given compounds, categorized as 0 through \textcolor{red}{\{$cluster\_number$\}}.
\\
input: ...
\end{tcolorbox}

\textbf{Data curation}:
The RT annotation prompt is employed to guide the LLM-RT model in annotating the RT during the ST and Annotate process.
The term `task' in the prompt can be dynamically substituted with specific tasks such as forward reaction prediction, reverse reaction synthesis, or reagent prediction, adapting the prompt to various contexts of CRPs.
Meanwhile, `cluster\_number' corresponds to the number of clusters $N$ identified in the Self-feedback Cluster process.
Utilizing the trained and frozen LLM-RT, we annotate the RTs for the validation and testing datasets using RT annotation prompts and input information.
The ${RT}$ performs annotation on the input data of ${D_{test}}$ using the $Annotate$ process facilitated by the LLM-RT. 
\begin{equation}
{RT}_{D_{test}} = Annotate(input_{D_{test}})
\end{equation}

\subsection{Adaptive Knowledge Injection}

The application of prompt learning to infuse extracted knowledge priors into our models demonstrates how this approach boosts the predictive accuracy of chemical reactions.
\begin{tcolorbox}[sharp corners, boxrule=0.5pt, colback=white]
\textbf{Instruction Templates:}

\textit{forward:}
\begin{itemize}
    \item ``Please suggest a potential product based on the given reactants and reagents.''
    \item ``Please provide a feasible product that could be formed using the given reactants and reagents.''
    \item ``Based on the given reactants and reagents, what product could potentially be produced?''
    \item \ldots
\end{itemize}

\textit{retrosynthesis:}
\begin{itemize}
    \item ``Provided the product below, propose some possible reactants that could have been used in the reaction.''
    \item ``Please suggest potential reactants used in the synthesis of the provided product.''
    \item ``Given these product, can you propose the corresponding reactants?''
    \item \ldots
\end{itemize}

\textit{reagent:}
\begin{itemize}
    \item ``Based on the given chemical reaction, can you propose some reagents that might have been utilized?''
    \item ``Can you provide potential reagents for the following chemical reaction?''
    \item ``Please suggest some possible reagents that could have been used in the following chemical reaction.''
    \item \ldots
\end{itemize}

\end{tcolorbox}

\begin{figure*}[!t]
\centering
\includegraphics[width=0.9\textwidth]{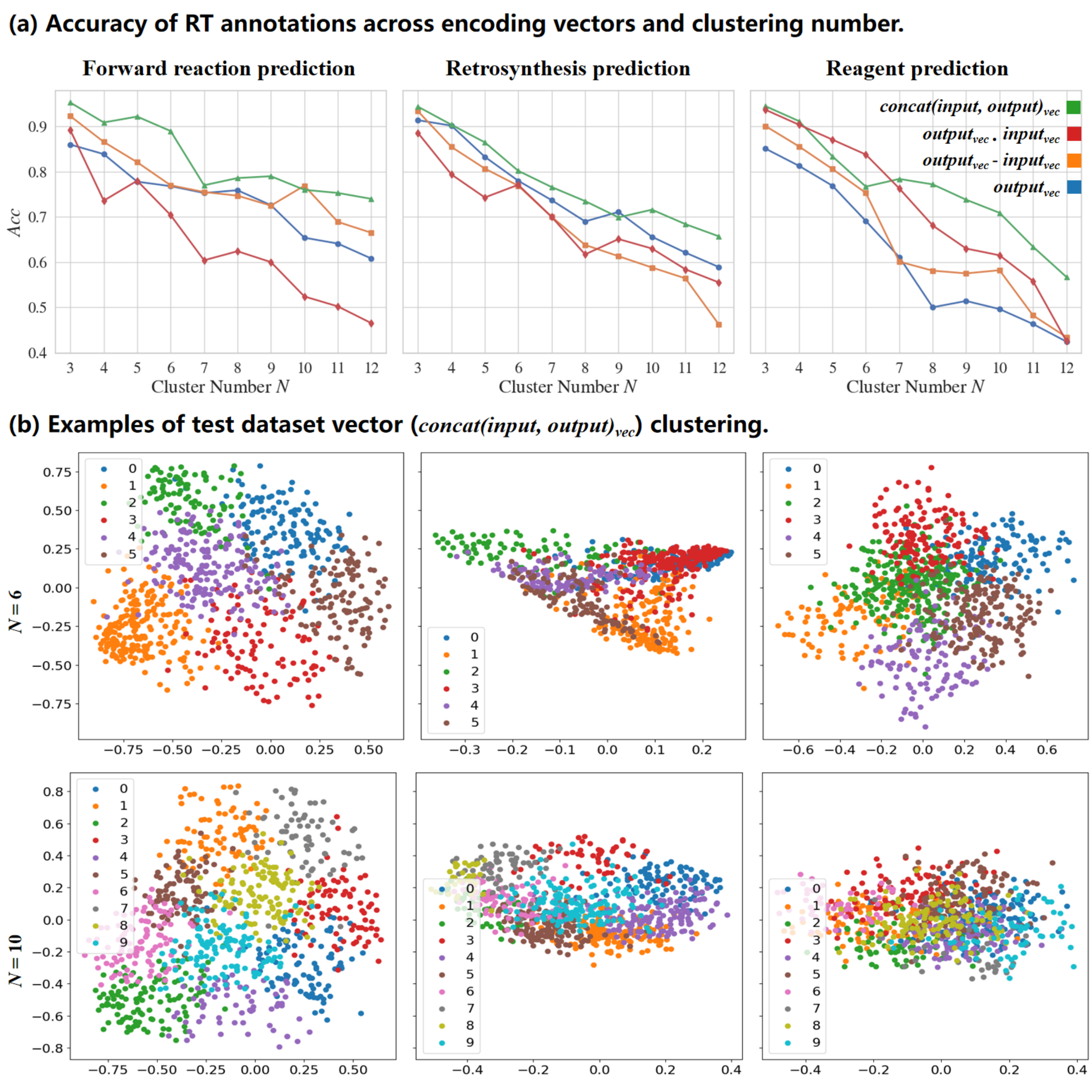}
\caption{\textbf{Performance of encoding vector self-feedback annotation and clustering}. \textbf{(a) Accuracy of RT annotations across encoding vectors and clustering number}, we compare the annotation accuracy $Acc$ among four encoding methods alongside reasonable Cluster Numbers $N$.
The results indicate that the encoding method using ($concat(input, output)_{vec}$) yields the best performance.
\textbf{(b) The test dataset vector ($concat(input, output)_{vec}$) clustering}, with the $N$ set to 6 and $N$ set to 10, test dataset vectors are reduced to two dimensions via a linear layer to display the clustering outcome.}
\label{fig: clustering}
\end{figure*}

\textbf{Adaptive instruction}:
To address the constraints of static templates and improve model generalization, we introduce an adaptive selector for template selection.
We establish an instruction template library, allocating 12 distinct templates per task, for a total of 36 templates.
During the embedding process, the $input_{i}$ from dataset $D$ and all corresponding templates from the task-specific list in the instruction template library are embedded via the LLM-CRP model.
The adaptive selector then evaluates adaptability through vector differences between the input embedding and each template embedding within the library. For each batch, this process entails matching a single input with multiple instructions to determine the best fit based on adaptability scores. The template that exhibits the highest adaptability with the input is chosen to facilitate precise knowledge injection.
\begin{equation}
Adaptability_{i,j} = -\| Emb_{input} - Emb_{instruction} \|_2
\end{equation}
\begin{equation}
IA_i = \arg\min_j(Adaptability_{i,j})
\end{equation}
The $Emb_{input}$ is the embedding of the $i^{th}$ input, transforming it into a vector representation that captures its semantic properties within the model's learned feature space.
Similarly, $Emb_{instruction}$ calculates the embedding of the $j^{th}$ instruction template.
The $Adaptability_{i,j}$ quantifies the similarity score between the $i^{th}$ input and the $j^{th}$ instruction template by maximizing the negative Euclidean distance, thus indicating higher relevance when the value is larger.
The adaptive selector then determines the most suitable instruction for the $i^{th}$ input by selecting the template that maximizes the $Adaptability_{i,j}$ score.
This process effectively identifies the instruction that best aligns with the input, optimizing the knowledge injection based on the adaptive selector's analysis.

\begin{table*}[!h]
\centering
\caption{\textbf{Performance comparison of various models on the reactions task}.
The results for Text+Chem T5 and nach0 are taken from their respective publications, whereas T5, MolT5, and Text+Chem T5 (finetune) represent performance after finetuning on a specific dataset.
The abbreviation $RT$ stands for reaction type, $N$ denotes the number of clusters, and $IA$ signifies Instruction-adaptive. Using Text+Chem T5 (finetune) as a baseline, ours($RT$+$IA$, $N$=10) shows improved $EM_{score}$ performance, particularly in retrosynthesis and reagent prediction tasks.}
\small
\resizebox{0.9\linewidth}{!}{%
\begin{tabular}{@{}c|ccccccc@{}}
\toprule
\textbf{Task} & \textbf{Model} & \textbf{$Bleu_{score}$} & \textbf{$Meteor_{score}$} & \textbf{$EM_{score}$} & \textbf{$Similarity_{score}$} & \textbf{$Validity_{score}$} & \textbf{$improve$} \\
\midrule
\multirow{8}{*}{\textbf{Forward}} & Text+Chem T5 & \textemdash & \textemdash & 0.594 & \textemdash & \textemdash & \textemdash \\
                         & nach0 & \textemdash & \textemdash & 0.890 & \textemdash & \textemdash & \textemdash \\
                         & T5 & 0.986 & 0.989 & 0.926 & 0.984 & 0.992 & \textemdash \\
                         & MolT5 & 0.986 & 0.988 & 0.897 & 0.978 & 0.992 & \textemdash \\
                         & Text+Chem T5 (finetune) & 0.988 & 0.991& 0.932 & \textbf{0.985} & \textbf{\textcolor{red}{0.999}} & \textemdash \\
\cmidrule{2-8}
                         & ours($RT$,$N$=10) & \textbf{\textcolor{red}{0.991}} & 0.991 & \textbf{0.937} & 0.984 & \textbf{0.997} & 0.5\% \\
                         & \textbf{ours($RT$+$IA$, $N$=10)} & \textbf{\textcolor{red}{0.991}} & \textbf{\textcolor{red}{0.993}} & \textbf{\textcolor{red}{0.945}} & \textbf{\textcolor{red}{0.986}} & \textbf{0.997} & 1.4\% \\
                         & ours($RT$+$IA$,$N$=6) & \textbf{0.989} & \textbf{0.992} & 0.930 & 0.984 & 0.995 & -0.2\% \\
\midrule
\multirow{8}{*}{\textbf{Retrosynthesis}} & Text+Chem T5 & \textemdash & \textemdash & 0.372 & \textemdash & \textemdash & \textemdash \\
                                & nach0 & \textemdash & \textemdash & 0.390 & \textemdash & \textemdash & \textemdash \\
                                & T5 & 0.920 & 0.921 & 0.649 & 0.855 & 0.989 & \textemdash \\
                                & MolT5 & 0.918 & 0.920 & 0.637 & 0.846 & 0.987 & \textemdash \\
                                & Text+Chem T5 (finetune) & 0.926 & 0.929 & 0.663 & 0.858 & \textbf{0.997} & \textemdash \\
\cmidrule{2-8}
                                & ours($RT$,$N$=10) & \textbf{0.941} & \textbf{0.947} & \textbf{0.749} & \textbf{0.895} & 0.996 & 13.0\% \\
                                & \textbf{ours($RT$+$IA$, $N$=10)} & \textcolor{red}{\textbf{0.944}} & \textcolor{red}{\textbf{0.950}} & \textcolor{red}{\textbf{0.757}} & \textcolor{red}{\textbf{0.905}} & 0.994 & 14.2\% \\
                                & ours($RT$+$IA$,$N$=6) & 0.920 & 0.921 & 0.654 & 0.848 & \textcolor{red}{\textbf{0.998}} & -1.4\% \\
\midrule
\multirow{7}{*}{\textbf{Reagent}} & nach0 & \textemdash & \textemdash & 0.140 & \textemdash & \textemdash & \textemdash\\
                         & T5 & 0.506 & 0.654 & 0.168 & 0.548 & 0.998 & \textemdash\\
                         & MolT5 & 0.515 & 0.660 & 0.178 & 0.559 & 0.997 & \textemdash\\
                         & Text+Chem T5 (finetune) & 0.482 & 0.657 & 0.163 & 0.571 & 0.996 & \textemdash\\
\cmidrule{2-8}
                         & ours($RT$,$N$=10) & \textbf{0.589} & \textbf{0.728} & \textbf{0.273} & \textbf{0.640} & \textbf{0.999} & 67.4\%\\
                         & \textbf{ours($RT$+$IA$, $N$=10)} & \textcolor{red}{\textbf{0.617}} & \textcolor{red}{\textbf{0.744}} & \textcolor{red}{\textbf{0.284}} & \textcolor{red}{\textbf{0.649}} & \textcolor{red}{\textbf{1.000}} & 74.2\%\\
                         & ours($RT$+$IA$,$N$=6) & 0.499 & 0.665 & 0.175 & 0.587 & \textbf{0.999} & 7.4\%\\
\bottomrule
\end{tabular}%
}

\label{table:performance_metrics}
\end{table*}

\textbf{Enhanced prompt}:
The selected adaptive instruction is combined with the corresponding RT for the current input to form an $prompt_{enhanced}$.
This enhanced prompt is then amalgamated with the $input$ and injected into LLM-CRP, aiming to refine the model's response to the input based on the tailored guidance.
\begin{equation}
prompt_{enhanced\_i} = fuse(IA_i, RT)
\end{equation}
\begin{equation}
(input_i, prompt_{enhanced\_i}) \rightarrow (\mathcal{R}', \mathcal{P}', \mathcal{G})
\end{equation}
The $prompt_{enhanced\_i}$ represents the process of creating an enhanced prompt for the $i^{th}$ input by fusing the adaptively selected instruction $instruction_{adaptive\_i}$ with the corresponding $RT$.
This fusion process ($fuse$) generates a $prompt_{enhanced}$ that is specifically tailored to both the context of the input and the instructional guidance deemed most appropriate by the adaptive selection mechanism.
Subsequently, the pair $(input_i, prompt_{enhanced\_i})$ is fed into LLM-CRP.
The model's output, represented as $(\mathcal{R}', \mathcal{P}', \mathcal{G})$, encompasses the predictions of reactants \(\mathcal{R}'\), products \(\mathcal{P}'\), and the necessary reagents \(\mathcal{G}\).


\begin{table*}[!h]
\centering
\caption{\textbf{Multi-task performance summary}.
T5, MolT5, and Text+Chem T5 models are finetuned with aggregated data from three tasks.
The average single-task results of Text+Chem T5 are compiled as Text+Chem T5 (avg-tasks), serving as the baseline.
The integrated model with $prompt_{enhanced}$ exhibits a \textbf{14.9\%} improvement over this baseline.
Joint training on multiple tasks tends to decrease performance, yet there is a \textbf{17.8\%} enhancement over Text+Chem T5 (finetune), highlighting the role of $RT$ and $IA$ in multi-task collaboration.
}
\small
\resizebox{0.9\linewidth}{!}{%
\begin{tabular}{@{}c|cccccc@{}}
\toprule
\textbf{Model} & \textbf{$Bleu_{score}$} & \textbf{$Meteor_{score}$} & \textbf{$EM_{score}$} & \textbf{$Similarity_{score}$} & \textbf{$Validity_{score}$} & \textbf{$improve$} \\
\midrule
T5 & 0.825 & 0.854 & 0.556 & 0.790 & 0.992 & \textemdash \\
MolT5 & \textbf{0.837} & \textbf{0.859} & \textbf{0.586} & 0.797 & 0.996 & \textemdash \\
Text+Chem T5 (finetune) & 0.822 & 0.857 & 0.572 & 0.797 & 0.995 & \textemdash \\
Text+Chem T5 (avg-tasks) & 0.799 & \textbf{0.859} & \textbf{0.586} & \textbf{0.805} & \textbf{0.997} & \textemdash \\
\midrule
ours($RT$+$IA$,$N$=10) & \textcolor{red}{\textbf{0.879}} & \textcolor{red}{\textbf{0.901}} & \textcolor{red}{\textbf{0.674}} & \textcolor{red}{\textbf{0.854}} & \textcolor{red}{\textbf{0.998}} & 14.9\% \\
\bottomrule
\end{tabular}%
}
\label{table:performance_metrics_reactions}
\end{table*}

\section{Evaluation and Results}
\label{sec:eva}
In this chapter, we subject our proposed methodologies to a rigorous evaluation aimed at addressing three fundamental research questions that guide our investigation.
For KI1, we set the number of clusters to range from 3 to 12 and tested four different embedding techniques, selecting the optimal clustering number and encoding method based on annotation accuracy.
Regarding KI2, we infuse the RT into the LLM using prompts and assess the effectiveness of adaptive prompt learning.
Finally, for KI3, we integrate chemical reaction data to perform comprehensive fine-tuning, achieving results from multi-task training.
\begin{itemize}
    \item KI1: How can we balance annotation accuracy and number of RTs in knowledge elicitation by LLMs?
    \item KI2: Can LLMs perform better through prompt-based knowledge infusion?
    \item KI3: Can a multi-task collaborative approach improve the performance of LLMs?
\end{itemize}

\subsection{Experimental Setup}

This subsection provides an overview of the training configurations and evaluation metrics.
We detail the model settings and hyperparameter values in model training.
In the knowledge extraction phase of our RT annotation experiments, we employ multi-class accuracy ($Acc$) as the primary evaluation metric.
This metric measures the proportion of correctly identified RTs among all predictions, providing a straightforward assessment of the model's performance in categorizing chemical reactions into their correct types.

\textbf{Training settings}:
Training employs Tesla V100-SXM2-32GB GPUs with CUDA 11.7 and PyTorch 2.0.0, leveraging AdamW for optimization.
Batch sizes vary from 24 to 48, with the patience of 2 epochs for early stopping to curb overfitting.
Training spans up to 40 epochs, using adaptive learning rates between 1e-4 and 1e-3 to finetune speed and stability.

\textbf{Evaluation metrics}: 
During the knowledge injection phase for CRPs, our evaluation strategy is more comprehensive, incorporating a blend of NLP evaluation metrics and compound generation assessment metrics.
We include BLEU scores ($Bleu_{score}$), gauging the linguistic similarity between the generated text and reference sequences, and METEOR scores ($Meteor_{score}$), offering a more nuanced evaluation by considering sentence structure.
Furthermore, to assess the chemical relevance and accuracy of the generated compounds, we introduce a similarity metric ($Similarity_{score}$), quantifying the resemblance between generated and target compounds.
The validity metric ($Validity_{score}$) ensures that every generated compound is chemically valid.   The exact match score ($EM_{score}$) disregards the sequence in which compounds are generated, focusing on the presence of correct chemical entities.
These metrics provide a comprehensive view of the model's capability in CRPs.

\subsection{Knowledge Elicitation}

To address KI1, this section starts by outlining the objectives of the analysis, emphasizing the importance of selecting optimal encoding vectors and the number of clusters $N$ for RT annotation accuracy.
We introduce four encoding methods evaluated in the study: direct output vector, output minus input vector, concatenated input-output vector, and the dot product of input and output vectors.
These vectors are encoded using LLM-RT Text+Chem T5.
Then, we explain the rationale behind exploring different cluster numbers, highlighting the hypothesis that the choice of encoding and clustering can significantly impact the annotation accuracy $Acc$.

\textbf{Encoding methods}:
We delve into a comparative analysis of $Acc$ using various encoding vectors across different task types, with a focus on a range of $N$.
As illustrated in Figure \ref{fig: clustering} (a), $Acc$ gradually declines within the adaptable range of $N$ from 3 to 12.
The findings underscore that the concatenated input-output vector ($concat(input, output)_{vec}$) consistently achieves the highest annotation accuracy among different tasks, maintaining over 70\% accuracy even when $N$ is set to 10.
It highlights the concatenated vector's capability to capture the nuances of chemical RTs.

\textbf{Clustering visualization and implications}:
We examine the clustering outcomes for the test dataset, focusing on the optimal encoding method ($concat(input, output)_{vec}$) for selected cluster numbers, $N=6$ and $N=10$, with visualizations depicted in Figure \ref{fig: clustering} (b).
The process involves reducing the high-dimensional encoded vectors to two dimensions for visualization, enabling us to observe the cluster distributions.
These visualizations reveal that chemical reactions for each task display distinct knowledge patterns.
Although the direct implications of these tasks remain unspecified, the identified patterns serve as crucial prompt information for generating reaction content.

\subsection{Knowledge Injection}

\begin{figure*}[!h]
\centering
\includegraphics[width=0.85\linewidth]{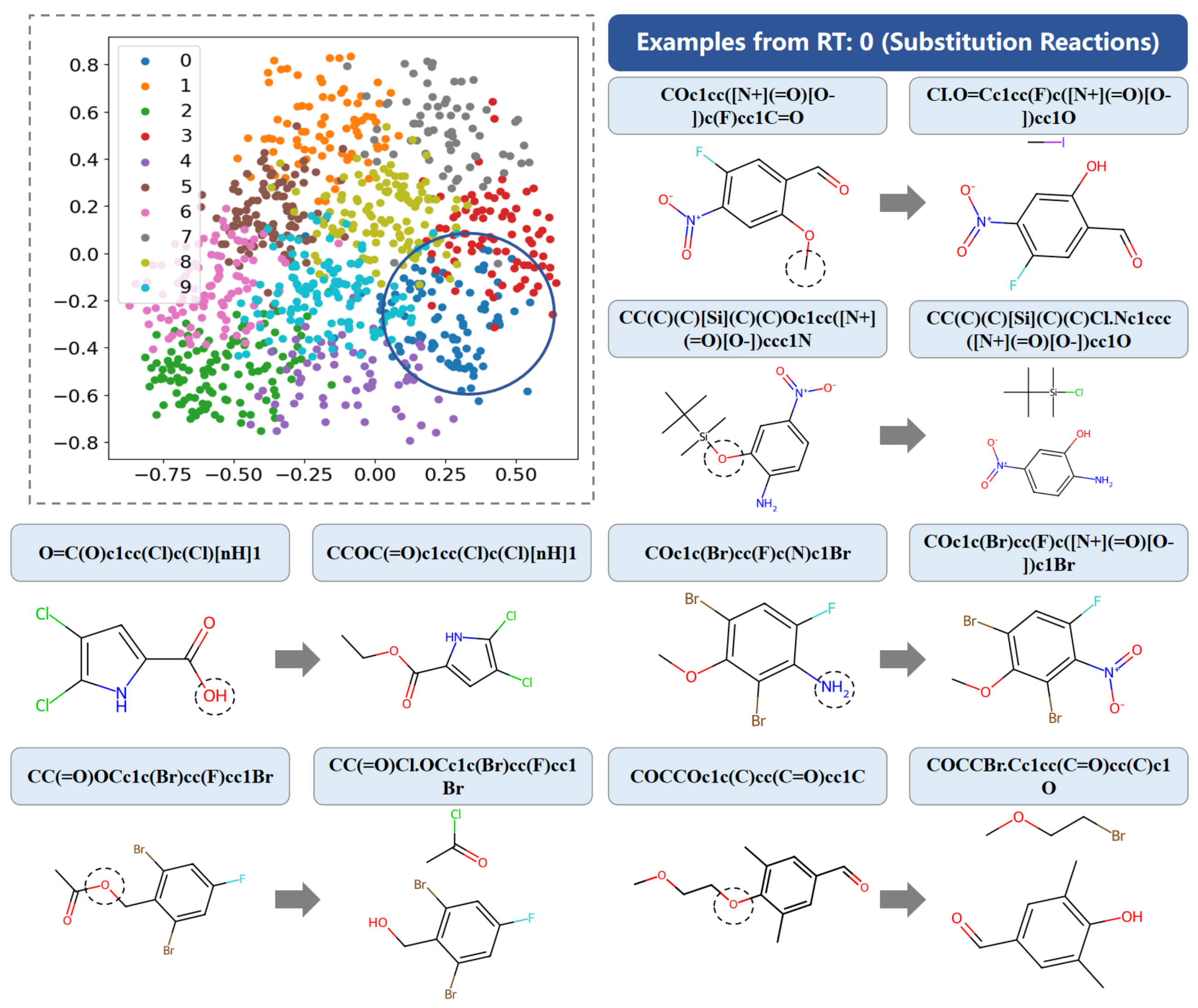}
\caption{\textbf{Case studies of RT annotation}.
To validate the practical significance of RT annotation, we filter through the \( concat(input, output)_{vec} \) vector with \( N=10 \) labeled results, focusing on samples with an RT label of 0.
The molecules in these instances transform simple atomic substitutions.
This analysis verifies the predominance of substitution reactions within these cases, demonstrating the real-world relevance of our RT annotation method.
}
\label{fig: RT_case_study}
\end{figure*}

This subsection assesses the effectiveness of integrating knowledge priors via prompt learning, directly responding to KI2 by comparing the predictive advantages gained through our knowledge-infused model against baseline approaches.
In our study, we utilize Text+Chem T5 as the LLM-CRP.
The adaptive selector chooses appropriate adaptive instructions, which are then fused with RT to create a prompt-enhanced input.
Subsequently, the input is fed into the LLM-CRP for training and testing.
The results are shown in Table \ref{table:performance_metrics}.

\textbf{RT integration}: Selecting an optimal $N$ and integrating $RT$ leads to significant performance enhancement, especially in retrosynthesis and reagent prediction, with improvements of 14.2\% and 74.2\%.

\textbf{Instruction adaptation}: This effect is especially pronounced in retrosynthesis and reagent prediction tasks, with almost a 10\% additional increase (13.0\% $\rightarrow$ 14.2\% and 67.1\% $\rightarrow$ 74.2\%)

\textbf{Cluster number}: When the cluster number $N$ is set too low, despite achieving high standard accuracy rates, the potential benefits of $RT$ knowledge injection may not be fully realized, and could inadvertently result in a decline in model performance.

\subsection{Multi-Task Reaction Prediction}

To tackle KI3, we evaluate the performance of our model on multi-task reaction prediction, highlighting how prompt-based knowledge injection influences the model’s ability to accurately predict various chemical reaction tasks.
In our study, we amalgamate datasets from three distinct tasks for fine-tuning models, including T5, MolT5, and Text+Chem T5.  Assuming each task is trained separately, we can establish an optimal performance baseline for each, denoted as Text+Chem T5 (avg-tasks).
The outcomes of this experiment are detailed in Table \ref{table:performance_metrics_reactions}.

\textbf{The side effects of multi-tasking}:
Based on the experimental results of Text+Chem T5 (avg-tasks) and Text+Chem T5 (finetune), we observe that direct integration and training across multiple tasks can lead to conflicting gradient updates and task interference, ultimately degrading the model's performance.
This phenomenon, often referred to as negative transfer, occurs when the optimization for one task adversely affects the learning of another, resulting in suboptimal performance across the board.

\textbf{Synergy of prompt-enhanced learning}:
The Prompt-Enhanced approach demonstrates a synergistic effect across multiple tasks, not only counteracting the side effects but also fostering synergy.
This approach results in a significant performance boost of 14.9\% over Text+Chem T5 (avg-tasks) due to improved task-specific adaptation and focused learning.
By guiding the model with task-specific prompts, we effectively mitigate the issues of task interference, enabling the model to leverage shared knowledge bases while honing in on the nuances of each task.
The enhanced task formulation provided by the prompts leads to more effective learning strategies and superior overall performance.


\section{Discussion}
\label{sec:res}

In our exploration, we've highlighted the innovative approach to RT annotation as a solution to the pervasive challenge of data scarcity in real-world scenarios.
Our methodology showcases the LLM's inherent ability to internalize and utilize latent knowledge, asserting the necessity of precise guidance to unlock its full potential.
Further, our analysis extends to the multi-task benefits derived from this guided learning process.
We aim to illuminate the critical insights and limitations encountered throughout our study.

\textbf{RT annotation significance}:
From Table \ref{table:performance_metrics} in Chapter \ref{sec:eva}, the practical effect of RT Annotation injection into language models is evident.
The model's performance can be improved with higher cluster numbers N, but this also introduces challenges with annotation accuracy.
Thus, finding a balance between annotation accuracy and the quantity of $N$ emerges as a focal point of this part.
More importantly, we scrutinize whether RTs annotated through self-feedback knowledge elicitation correspond to actual chemical RTs.
This alignment between automated annotations and human-understandable concepts can significantly propel the advancement of interpretability in LLMs.
Figure \ref{fig: RT_case_study} presents randomly selected instances of reactions with RT annotations labeled as 0, with the reaction sites indicated.
Most of these reactions are identified as substitution reactions, which confirms the practical significance of the knowledge patterns that our knowledge elicitation methodology extracts.

\textbf{Knowledge learning in LLMs}:
LLMs might intrinsically possess the capability to predict RTs.
The slight improvements in forward reaction predictions illuminate this ability of LLMs.
LLMs can also grasp an understanding of reaction mechanisms without explicit instruction.
By deconstructing the problem, we allow the models to gain prior knowledge, facilitating the planning of reaction pathways and simplifying the text generation task into more straightforward classification issues and less complex text generation tasks.
Consequently, this can enhance the overall task performance while also increasing the complexity of the process.
This approach to problem decomposition is also potentially applicable across various scientific domains, such as molecule design and lead optimization, suggesting a broad utility of this methodology in advancing research and understanding in multiple fields.

\textbf{Knowledge-enhanced multi-task synergy}:
We delve into the mechanisms underlying the significant uptick in model accuracy for multi-task learning facilitated by the use of enhanced prompts and knowledge injection.
The integration of contextually rich prompts and targeted knowledge snippets acts as a catalyst, fine-tuning the model's focus and understanding of each task.
This approach not only amplifies task-specific performance but also harmonizes the learning process across disparate tasks.
The prior knowledge acts as an anchor, grounding the model's learning process in real-world phenomena and relationships, thereby reducing the ambiguity inherent in complex tasks. 
For example, understanding the relationship between molecular structure and pharmacological activity in one task can enhance the model's ability to predict drug toxicity in another, as both tasks share underlying chemical knowledge.
Furthermore, this synergy underscores the potential of structured knowledge and task-specific prompts in augmenting the intrinsic multi-tasking capabilities of LLMs.

In future research, the adaptive algorithms can accurately determine the optimal number and encoding strategies for knowledge partners, thus avoiding the inefficient trial-and-error approach.
Exploring the potential of dynamic prompts that not only derive from a broader, more randomized pool but also retain the ability to guide specific tasks promises to improve model performance.
While the extraction and injection of RTs have enhanced interpretability to a degree, they fall short of revealing the model's exploration within the chemical space.
The development of tools for knowledge visualization and tracking would enable the pinpointing of how RTs guide the text generation process, underutilizing the potential of LLMs.

\section{Conclusion}
\label{sec:con}
In this study, we reconceptualize the task at hand as SLM and pioneer a data-curated, self-feedback knowledge elicitation method to identify knowledge partners, specifically RTs.
We then employ dynamic prompt learning to integrate this prior knowledge into LLMs, thereby enhancing accuracy in CRPs and across multiple-task CRPs.
This research sets a novel paradigm for knowledge elicitation within scientific domains and for the integration of knowledge priors, laying foundational groundwork for the advancement of SLM.


%



\ifCLASSOPTIONcompsoc
  \section*{Acknowledgments}
\else
  \section*{Acknowledgment}
\fi

This work is supported by grants from the National Natural Science Foundation of China (62172456 and 62372484), the Major Key Project of PCL PCL2021A13 and Peng Cheng Cloud-Brain.

\ifCLASSOPTIONcaptionsoff
  \newpage
\fi



\bibliographystyle{IEEEtran}
\bibliography{arxiv}
%



%

\begin{IEEEbiography}
[{\includegraphics[width=1in,height=1.25in,clip,keepaspectratio]{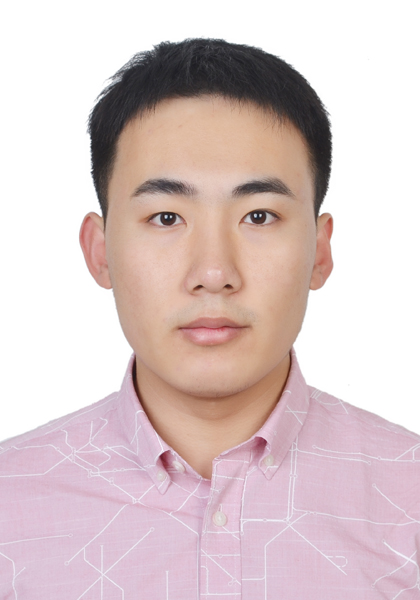}}]{Pengfei Liu}
is a Ph.D. candidate in the School of Computer Science and Engineering at Sun Yat-sen University, with a concurrent affiliation to Peng Cheng Laboratory. 
His research interests include large language models, multi-modal learning, molecular modeling and design, and visual analytics.
\end{IEEEbiography}

\begin{IEEEbiography}
[{\includegraphics[width=1in,height=1.25in,clip,keepaspectratio]{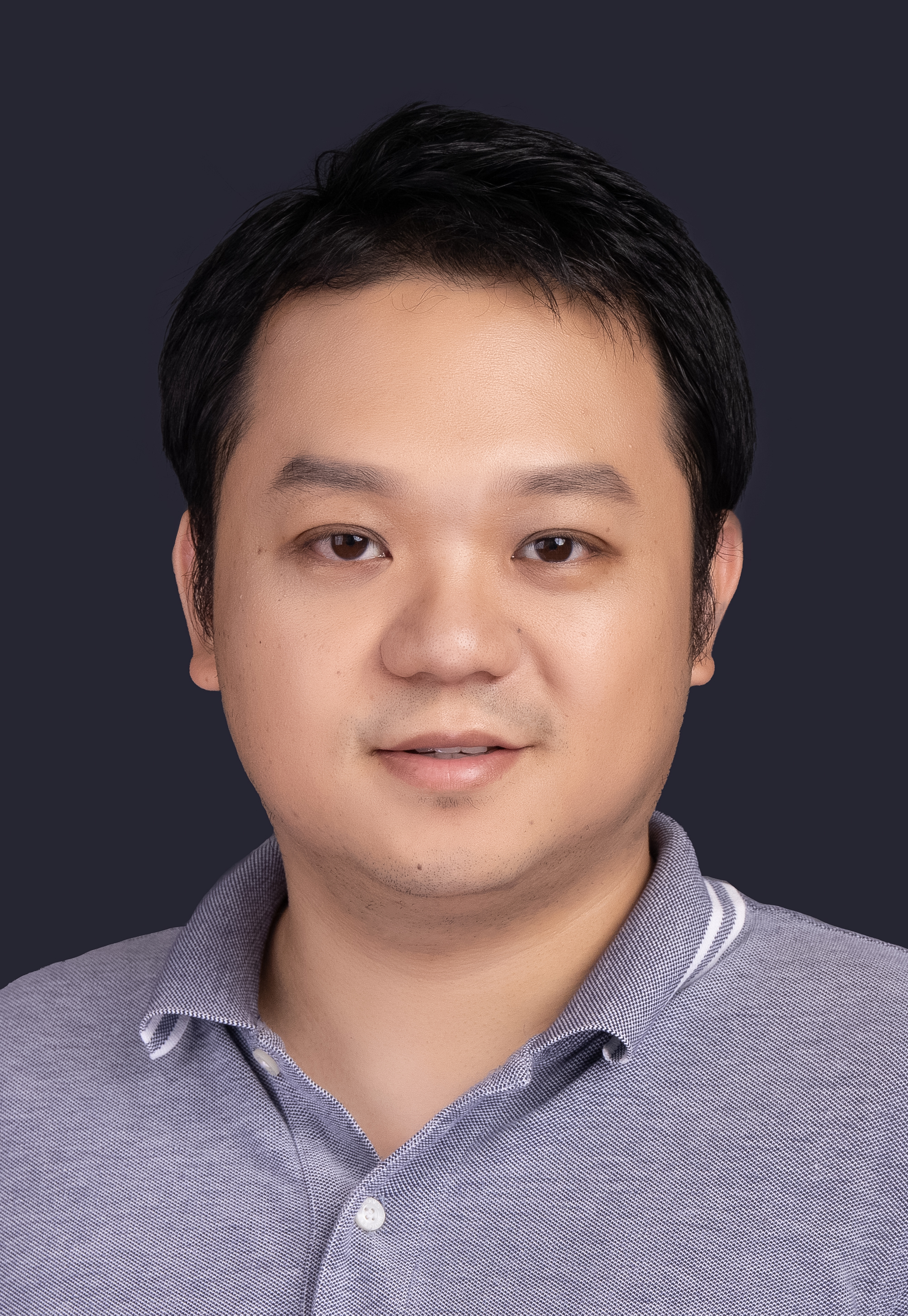}}]{Jun Tao} is an associate professor of computer science at Sun Yat-sen University and National Supercomputer Center in Guangzhou. He received a Ph.D. degree in computer science from Michigan Technological University in 2015. His research interest lies at the intersection of visualization, learning approaches, and science discoveries, with a focus on developing novel interactive techniques that combines human and machine intelligence to explore flow and scalar fields.
\end{IEEEbiography}

\begin{IEEEbiography}
[{\includegraphics[width=1in,height=1.25in,clip,keepaspectratio]{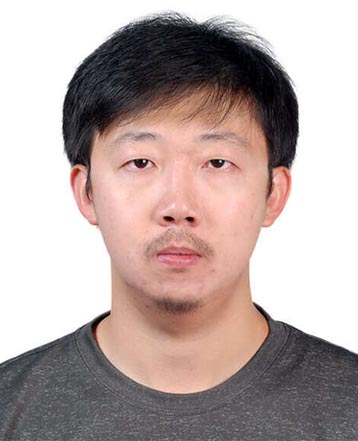}}]{Zhixiang Ren} is currently an Associate Research Scientist at Peng Cheng Laboratory and a PhD Supervisor at Southern University of Science and Technology in Shenzhen, China. He received his PhD from the University of New Mexico (Albuquerque, USA) in 2018. His research interests include AI for science and multi-modal large AI models, especially "AI+bioinformatics" and AI-aided drug design. He has published over 50 papers with more than 7,000 citations. He has been granted several patents and has led the development of one international standard and two industry standards in the field of intelligent computing. He is also an associated editor of "Frontiers in Big Data", "Big Data Mining and Analytics" and "CAAI Artificial Intelligence Research", as well as an experienced reviewer for multiple renowned journals such as Neural Networks and Pattern Recognition.
\end{IEEEbiography}







\end{document}